\PassOptionsToPackage{table}{xcolor}
\documentclass[11pt]{article}

\usepackage[preprint]{acl}

\usepackage{times}
\usepackage{latexsym}

\usepackage[T1]{fontenc}

\usepackage[utf8]{inputenc}

\usepackage{microtype}

\usepackage{inconsolata}

\usepackage{graphicx}
\usepackage{amsmath}
\usepackage{booktabs}

\title{Conditional Hypothesis Generation for LLM-Based Text Analysis with Researcher-Specified Covariates}

\author{Paiheng Xu, Jing Liu \and Wei Ai \\ 
        \\
        University of Maryland, College Park \\
        \small
        \texttt{\{paiheng,jliu28,aiwei\}@umd.edu}}

\usepackage{xspace}
\usepackage{tikz}
\newcommand{\abr}[1]{\textsc{#1}\xspace}
\usepackage{amsfonts}
\usepackage{amssymb}
\usepackage{booktabs}
\usepackage{tcolorbox}
\usepackage{listings}
\usepackage{array}
\usepackage{enumitem}
\usepackage{colortbl}
\usepackage{xcolor}

\newcommand{\llm}{\abr{llm}}

\newcommand{\sae}{\abr{sae}}
\newcommand{\lasso}{\abr{lasso}}
\newcommand{\congress}{\abr{congress}}
\newcommand{\ncte}{\abr{ncte}}
\newcommand{\remed}{\abr{remed}}
\newcommand{\css}{\abr{css}}

\definecolor{congressRepLight}{RGB}{247,221,221}
\definecolor{congressRepLighter}{RGB}{252,237,237}
\definecolor{congressDemLight}{RGB}{221,232,247}
\definecolor{congressDemLighter}{RGB}{236,243,252}
\definecolor{ncteHMLight}{RGB}{223,242,223}
\definecolor{ncteHMLighter}{RGB}{236,248,236}
\definecolor{ncteLowLight}{RGB}{247,221,221}
\definecolor{ncteLowLighter}{RGB}{252,237,237}

\newcommand{\congressRep}[1]{\cellcolor{congressRepLight}#1}
\newcommand{\congressRepAlt}[1]{\cellcolor{congressRepLighter}#1}
\newcommand{\congressDem}[1]{\cellcolor{congressDemLight}#1}
\newcommand{\congressDemAlt}[1]{\cellcolor{congressDemLighter}#1}
\newcommand{\ncteHM}[1]{\cellcolor{ncteHMLight}#1}
\newcommand{\ncteHMAlt}[1]{\cellcolor{ncteHMLighter}#1}
\newcommand{\ncteLow}[1]{\cellcolor{ncteLowLight}#1}
\newcommand{\ncteLowAlt}[1]{\cellcolor{ncteLowLighter}#1}

\begin{document}
\maketitle
\begin{abstract}

A core goal of computational social science is to discover interpretable differences in how language varies across outcomes of interest, such as political affiliation or instructional quality.
Recent \llm-based hypothesis generation methods describe such differences in natural language, but select for globally discriminative patterns without accounting for covariates that shape the data based on researchers' domain knowledge.
When covariates are ignored, selected patterns can reflect confounds rather than differences of substantive interest.
We introduce \emph{conditional} hypothesis generation, a framework that incorporates researcher-specified covariates to steer hypothesis discovery toward differences that hold within relevant subgroups.
Two challenges arise: the target subgroup may be underrepresented (\emph{stratum imbalance}), and the direction of a difference may reverse across subgroups (\emph{sign reversal}).
We propose two econometrics-inspired methods: one introduces feature--covariate interactions to detect sign reversals, and the other applies within-stratum demeaning and inverse-frequency reweighting to equalize underrepresented strata.
Synthetic experiments show each method outperforms global baselines in its targeted setting, and expert evaluation on two real-world datasets confirms that covariate-aware generation surfaces more useful hypotheses within relevant subgroups.

\end{abstract}

\section{Introduction}
\label{sec:intro}

A central goal of computational social science (\css) is to understand how text relates to variables such as political affiliation, instructional quality, or social media engagement.
Rather than predicting these outcomes, researchers seek interpretable hypotheses---natural-language descriptions of how textual patterns differ across outcome values---that can guide further investigation~\citep{grimmer2013text,card2019accelerating,grimmer2022text}.

Recent \llm-based methods support this form of analysis by sampling labeled examples and prompting an \llm to propose natural-language hypotheses that characterize textual patterns associated with different outcome values~\citep{zhong2022describing,zhong2023goal,zhong2024nlparam,zhou2024hypothesis,movva2025sparse}.
These methods typically select hypotheses by how well they discriminate between outcome groups.

However, global discrimination can be misleading.
A globally discriminative feature may reflect a confound rather than a difference of substantive interest,
a persistent concern in text-as-data research~\citep{grimmer2013text,gentzkow2019measuring,grimmer2022text}.
For example, \citet{taddy2013multinomial} shows that national-park language appears as a party-predictive feature because public lands are unevenly distributed across states,
though national park issues are not inherently partisan.
The challenge is to steer discovery toward differences that hold within conditions researchers care about---and to let researchers specify those conditions.

We introduce \emph{conditional} hypothesis generation, a framework for generating hypotheses that are discriminative within researcher-specified covariate strata.
Covariates---such as policy area, time period, or classroom environment---encode the domain knowledge that researchers bring to text analysis:
they define the conditions under which differences should be examined, without requiring the hypothesis itself to be known in advance.

Conditioning on covariates introduces two statistical challenges~\citep{simpson1951interpretation,gail1985testing}.
The target stratum may be underrepresented, so that its signal is dominated by larger strata (\emph{stratum imbalance}); or the direction of a difference may reverse across strata (\emph{sign reversal}), so that global aggregation cancels the conditional pattern.

We build on~\citet{movva2025sparse}, which maps documents to monosemantic Sparse Autoencoder (\sae) features and selects discriminative features via \lasso.
Because the \sae features are fixed before statistical selection, covariates can be incorporated directly into the selection step.
Drawing on econometrics, we propose two complementary methods.
\emph{Interaction-lasso} augments the feature space with feature--covariate interactions, allowing a feature to be selected when it is discriminative within a single stratum, even if its global effect is zero.
\emph{Demeaned-reweighted-lasso} residualizes features and outcomes within covariate strata to isolate within-stratum variation, and applies inverse-frequency weighting so that underrepresented strata contribute comparably to feature selection.

In synthetic evaluations with known ground-truth hypotheses and covariate structure, demeaned-reweighted-lasso outperforms global baselines and approaches oracle performance across imbalance levels, while interaction-lasso is the only method that recovers differences under sign reversal.

We validate on two real-world datasets:
\congress, a longstanding testbed for political language~\citep{gentzkow2010drives,grimmer2021machine}, and \ncte, a dataset of math classroom transcripts with rich annotations of instructional quality~\citep{demszky2023ncte,hill2008mathematical,pianta2012classroom}.
Expert evaluations show that covariate-aware selection surfaces hypotheses that domain experts rate as more useful than those unique to the global baseline.

Our contributions are:
(1) We formalize conditional hypothesis generation for text analysis with researcher-specified covariates.
(2) We introduce two complementary covariate-aware methods, each targeting a distinct statistical challenge (i.e., stratum imbalance and sign reversal).
(3) We design controlled synthetic evaluations covering these two challenges.
(4) Expert evaluation on two real-world datasets shows that covariate-aware methods guide discovery toward hypotheses that domain experts find more useful.

\section{Preliminaries}
\label{sec:prelim}

\subsection{Task Formulation}
\label{sec:task}

We consider a dataset $\{(x_i, y_i)\}_{i \in [N]}$, where $x_i$ is a text document and $y_i$ is a target variable.
In the simplest case, $y_i$ indicates group membership of the document.
The goal of \emph{hypothesis generation} is to produce a set of natural-language statements $\mathcal{H}$ that characterize how $y_i$ varies with text content, where $|\mathcal{H}|$ is a pre-specified number of hypotheses to generate.
Each text is associated with a set of covariates $\mathcal{V}$, which may include metadata (e.g., time period, author demographics) or attributes inferable from text (e.g., topic).
In our setting, a researcher may specify a subset $\mathbf{C}$ of these covariates based on domain knowledge to guide hypothesis generation within relevant covariate strata.

\subsection{\llm-Based Hypothesis Generation}
\label{sec:background}

\llm-based hypothesis generation for comparing texts typically follows a sample-and-propose pattern: sample a small set of labeled examples, then ask an \llm to induce patterns and propose natural-language hypotheses from those examples.
Existing methods differ in their sampling strategies.
\citet{zhong2022describing,zhong2023goal} train a classification model to randomly sample the most discriminative texts.
\citet{zhou2024hypothesis} iteratively generate hypotheses from examples that the current candidate list mis-classifies.
These methods then filter candidate hypotheses by their discriminative power.

\citet{zhong2024nlparam,movva2025sparse} instead use statistical models to select the most predictive features over all $N$ samples,
then call an \llm to describe each selected feature from a sample of feature-related texts.
\citet{movva2025sparse}, the current state of the art, train a Sparse Autoencoder (\sae) on text embeddings to obtain a fixed set of interpretable features, then select those most predictive of $y$ via \lasso, as detailed in \S\ref{sec:sae}.
We build on this feature-selection view because it makes incorporating covariates $\mathbf{C}$ well-defined.

\subsection{Hypothesis Generation with \sae}
\label{sec:sae}

A \sae encourages each input to be reconstructed from only a small active subset of features \citep{cunningham2023sparse}.
This training objective tends to produce monosemantic features, where each dimension captures a coherent, interpretable concept. 
\citet{movva2025sparse} exploit this property with three steps:
\textbf{(1) SAE encoding}: a trained \sae maps each document embedding to a sparse vector of feature activations.
Each dimension in this vector corresponds to one learned \sae feature.
\textbf{(2) Feature selection}: a statistical model (e.g., \lasso) selects $|\mathcal{H}|$ features most predictive of $y$.
\textbf{(3) \llm interpretation}: an \llm describes the texts that contrast most strongly activating input documents (positive examples) vs. zero-activating input documents (negative examples) for each selected feature as a natural-language hypothesis; Table~\ref{tab:sae_interpretation_prompt} shows the prompt.
Step~2 accepts any statistical model, making it straightforward to condition feature selection on covariates $\mathbf{C}$, as we show next.

\section{Method}
\label{sec:method}

We define the goal of \emph{conditional} hypothesis generation as discovering group differences that are discriminative within strata of $\mathbf{C}$,
which brings two statistical challenges~\citep{simpson1951interpretation,blyth1972simpson,gail1985testing}:
(1) \textbf{stratum imbalance}: the relevant covariate stratum may be rare in the corpora, the targeted difference is dominated by irrelevant differences in other strata.
(2) \textbf{sign reversal}: the direction of the group difference changes across strata, so the global difference can cancel or misrepresent the targeted differences.
A classic example is Simpson's paradox~\citep{simpson1951interpretation}.
The targeted hypothesis can only be reliably generated with within-stratum comparison.
We adapt two well-studied econometric tools to \sae feature selection: interaction models and group fixed effects (implemented via within-stratum demeaning).
Both subsume the global case: features that are discriminative across all strata remain selected alongside stratum-specific ones.

\paragraph{Setup.}
Let $Z \in \mathbb{R}^{N \times M}$ denote \sae activation matrix, where $z_{i,m}$ is the activation of feature $m$ for document $i$ and $M$ is the number of \sae features.
Let $\mathbf{C} = [C_1,\ldots,C_P] \in \mathbb{R}^{N \times P}$ denote the covariate matrix, where $C_p$ is the $p$th covariate column and $P$ is the number of covariates.
For simplicity, we focus on binary corpus comparison ($y \in \{0,1\}^N$, encoding corpus $A$ or $B$).
The baseline \lasso fits L1-regularized logistic regression:
\begin{equation}
\label{eq:lasso}
  \hat{\beta} = \arg\min_\beta \ell(y, Z\beta) + \lambda \|\beta\|_1,
\end{equation}
and selects the top $|\mathcal{H}|$ features ranked by $|\hat{\beta}_m|$, capturing only globally discriminative features.

\subsection{Interaction-lasso}
\label{sec:interaction_lasso}
A standard approach for modeling covariate-specific effects in regression is to include \emph{interaction terms} (products of two variables)~\citep{angrist2009mostly}.
When an \sae feature is interacted with a covariate, the feature's coefficient can vary with that covariate, capturing within-stratum differences.
We apply this idea by augmenting the \sae activation matrix with the covariates $\mathbf{C}$ and feature--covariate interaction blocks $Z \odot C_p$ for each covariate $p \in [P]$, where $Z \odot C_p$ denotes row-wise multiplication of all $M$ \sae features by covariate $C_p$.
Let $\eta(\beta,\delta,\gamma)=Z\beta+\mathbf{C}\delta+\sum_{p=1}^P (Z \odot C_p)\gamma_p$.
We fit \lasso on the full augmented space:
\begin{equation}
  \hat{\beta}, \hat{\delta}, \hat{\gamma}
  = \arg\min_{\beta,\delta,\gamma}
  \ell\!\left(y, \eta(\beta,\delta,\gamma)\right)
  + \lambda \|(\beta,\delta,\gamma)\|_1.
\end{equation}
Here $\beta$ are \sae feature main effects, $\delta$ are covariate main effects, and $\gamma_p$ are interaction effects for covariate $p$.
We rank features by $\max(|\hat{\beta}_m|, \max_{p \in [P]} |\hat{\gamma}_{p,m}|)$, and select the top $|\mathcal{H}|$ features.
A feature qualifies if it is discriminative globally (large $|\hat{\beta}_m|$) or through any covariate-specific interaction (large $|\hat{\gamma}_{p,m}|$ for some $p$), directly implementing the stated goal.
The covariate main effects $\delta$ are included as nuisance controls and are not ranked as potential hypotheses.

However, interaction-lasso faces two practical limitations.
First, the feature space substantially expands by including $M{\times}P$ interaction features,
which increases computational cost and can make feature selection less stable in high dimensions~\citep{bien2013lasso}.
Second, \sae features are inherently sparse, so each interaction term $Z {\odot} C_p$ is doubly sparse: nonzero only for samples with $C_{i,p} {\neq} 0$ and an active feature.
When a covariate is rare, these terms become nearly all-zero, making the $\hat{\gamma}_{p,m}$ estimates noisy~\citep{crump2009dealing}.
We validate this empirically in \S\ref{sec:syn_results}.

\subsection{Demeaned-Reweighted-Lasso}
\label{sec:demeaned_lasso}
To overcome the practical limitations, we adapt another standard econometrics technique called demeaning,
which subtracts group means from observations to remove between-group variation and isolate within-group differences~\citep{angrist2009mostly}.
By the Frisch--Waugh--Lovell theorem~\citep{lovell1963seasonal}, this is analogous to including group fixed effects in the regression.
Specifically, we group documents by their observed covariate values, residualize both \sae activations $Z$ and $y$ within each group, then run \lasso on the residuals.
Unlike interaction-lasso, demeaning operates on the original $M$ features without variable expansion, and estimates each feature's within-stratum difference from all $N$ samples rather than sparse interaction terms, avoiding both limitations above.

Formally, let $g_i$ denote the covariate stratum of document $i$, and let $\bar{z}_m^{(g_i)}$ and $\bar{y}^{(g_i)}$ be the mean feature activation and mean label within that stratum.
We compute $\tilde{z}_{i,m} = z_{i,m} - \bar{z}_m^{(g_i)}$ and $\tilde{y}_i = y_i - \bar{y}^{(g_i)}$.
The \textbf{demeaned-lasso} then selects features by running \lasso on $(\tilde{Z}, \tilde{y})$.
The $\tilde{\beta}_m$ estimates capture the within-stratum difference associated with feature $m$, partialling out per-stratum mean shifts in both the outcome and the feature activations.

Additionally, to address stratum imbalance, we propose \textbf{demeaned-reweighted-lasso}: set sample weights $w_i$ proportional to the inverse frequency of stratum $g_i$
and select features by the weighted lasso on $(\tilde{Z}, \tilde{y}, w)$.
Reweighting equalizes stratum contributions to feature selection, allowing features from underrepresented target strata to be selected when their within-stratum differences are strong.

Demeaning-based methods assume \textbf{no qualitative interaction (NQI)}: the targeted difference has a consistent sign across strata~\citep{gail1985testing}.
Under NQI, the covariates are treated as nuisance fixed effects rather than variables that change the direction of the text--outcome relationship.
In the sign reversal scenario, NQI is violated and interaction-lasso should be used instead.
We also include \textbf{demeaned-lasso} to isolate the contribution of the reweighting step.


\section{Experiments on Synthetic Datasets}
\label{sec:experiments}

\subsection{Synthetic Dataset Construction}
\label{sec:dataset}

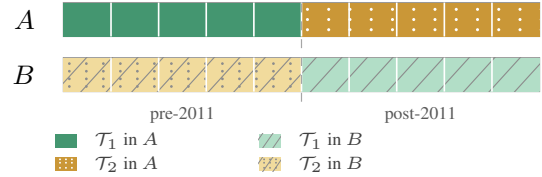
\begin{figure}[t]
\centering
\resizebox{\columnwidth}{!}{%
\begin{tikzpicture}[
    x=1cm,
    y=1cm,
    every node/.style={font=\scriptsize},
    corpuslabel/.style={font=\small\bfseries, anchor=east},
    paneltitle/.style={font=\scriptsize\bfseries, anchor=west},
    panelsubtitle/.style={font=\tiny, align=left, anchor=west, text=black!75},
    legendtext/.style={font=\tiny, anchor=west, text=black!75}
]
\definecolor{synthSoc}{RGB}{63,127,179}
\definecolor{synthSocLight}{RGB}{177,211,232}
\definecolor{synthGov}{RGB}{202,111,73}
\definecolor{synthGovLight}{RGB}{240,183,148}
\definecolor{synthOne}{RGB}{68,150,112}
\definecolor{synthOneLight}{RGB}{178,222,199}
\definecolor{synthTwo}{RGB}{202,151,55}
\definecolor{synthTwoLight}{RGB}{241,219,157}

\node[paneltitle] at (0,5.55) {Scenario 1: stratum imbalance};
\node[panelsubtitle, text width=6.6cm] at (0,5.22)
    {Many global differences exist, but covariates point to $\mathcal{T}_{soc}$. $r{=}0.35$};

\node[corpuslabel] at (0.85,4.58) {$A$};
\draw[black!40] (1.05,4.38) rectangle (6.65,4.78);
\fill[synthSoc] (1.05,4.38) rectangle (3.01,4.78);
\fill[synthGov] (3.01,4.38) rectangle (6.65,4.78);
\begin{scope}
    \clip (3.01,4.38) rectangle (6.65,4.78);
    \foreach \x in {3.10,3.35,...,6.70}
        \foreach \y in {4.46,4.58,4.70}
            \fill[white] (\x,\y) circle[radius=.018];
\end{scope}
\foreach \x in {1.70,2.36,3.01,4.22,5.44}
    \draw[white, line width=.55pt] (\x,4.38) -- (\x,4.78);
\node[font=\tiny\bfseries, text=white] at (2.03,4.58) {$r$};

\node[corpuslabel] at (0.85,3.93) {$B$};
\draw[black!40] (1.05,3.73) rectangle (6.65,4.13);
\fill[synthSocLight] (1.05,3.73) rectangle (4.69,4.13);
\fill[synthGovLight] (4.69,3.73) rectangle (6.65,4.13);
\begin{scope}
    \clip (1.05,3.73) rectangle (6.65,4.13);
    \foreach \x in {0.55,0.85,...,6.65}
        \draw[black!45, line width=.38pt] (\x,3.60) -- (\x+0.58,4.25);
\end{scope}
\begin{scope}
    \clip (4.69,3.73) rectangle (6.65,4.13);
    \foreach \x in {4.76,4.96,...,6.76}
        \foreach \y in {3.79,3.88,3.97,4.06}
            \fill[black!45] (\x,\y) circle[radius=.016];
\end{scope}
\foreach \x in {2.26,3.48,4.69,5.34,6.00}
    \draw[white, line width=.55pt] (\x,3.73) -- (\x,4.13);
\node[font=\tiny\bfseries, text=black!75] at (2.87,3.93) {$1-r$};

\draw[thick, synthSoc] (1.05,4.90) -- (3.01,4.90);
\draw[thick, synthSoc] (1.05,4.90) -- (1.05,4.83);
\draw[thick, synthSoc] (3.01,4.90) -- (3.01,4.83);
\node[font=\tiny, text=synthSoc] at (2.03,5.03) {$\mathcal{H}^*$: target strata};

\fill[synthSoc] (0.95,3.28) rectangle (1.20,3.44);
\node[legendtext] at (1.28,3.36) {$\mathcal{T}_{soc}$ in $A$};
\fill[synthSocLight] (3.35,3.28) rectangle (3.60,3.44);
\begin{scope}
    \clip (3.35,3.28) rectangle (3.60,3.44);
    \foreach \x in {3.25,3.37,...,3.60}
        \draw[black!45, line width=.3pt] (\x,3.24) -- (\x+0.18,3.50);
\end{scope}
\node[legendtext] at (3.68,3.36) {$\mathcal{T}_{soc}$ in $B$};
\fill[synthGov] (0.95,3.00) rectangle (1.20,3.16);
\foreach \x in {1.01,1.08,1.15}
    \foreach \y in {3.04,3.10,3.14}
        \fill[white] (\x,\y) circle[radius=.011];
\node[legendtext] at (1.28,3.08) {$\mathcal{T}_{gov}$ in $A$};
\fill[synthGovLight] (3.35,3.00) rectangle (3.60,3.16);
\begin{scope}
    \clip (3.35,3.00) rectangle (3.60,3.16);
    \foreach \x in {3.25,3.37,...,3.60}
        \draw[black!45, line width=.3pt] (\x,2.96) -- (\x+0.18,3.22);
\end{scope}
\foreach \x in {3.41,3.48,3.55}
    \foreach \y in {3.04,3.10,3.14}
        \fill[black!45] (\x,\y) circle[radius=.011];
\node[legendtext] at (3.68,3.08) {$\mathcal{T}_{gov}$ in $B$};

\node[paneltitle] at (0,2.57) {Scenario 2: sign reversal};
\node[panelsubtitle, text width=6.6cm] at (0,2.25)
    {Corpus composition reverses after 2011.};

\node[corpuslabel] at (0.85,1.60) {$A$};
\draw[black!40] (1.05,1.40) rectangle (6.65,1.80);
\fill[synthOne] (1.05,1.40) rectangle (3.85,1.80);
\fill[synthTwo] (3.85,1.40) rectangle (6.65,1.80);
\begin{scope}
    \clip (3.85,1.40) rectangle (6.65,1.80);
    \foreach \x in {3.92,4.17,...,6.67}
        \foreach \y in {1.48,1.60,1.72}
            \fill[white] (\x,\y) circle[radius=.018];
\end{scope}
\foreach \x in {1.61,2.17,2.73,3.29,3.85,4.41,4.97,5.53,6.09}
    \draw[white, line width=.55pt] (\x,1.40) -- (\x,1.80);

\node[corpuslabel] at (0.85,0.95) {$B$};
\draw[black!40] (1.05,0.75) rectangle (6.65,1.15);
\fill[synthTwoLight] (1.05,0.75) rectangle (3.85,1.15);
\fill[synthOneLight] (3.85,0.75) rectangle (6.65,1.15);
\begin{scope}
    \clip (1.05,0.75) rectangle (6.65,1.15);
    \foreach \x in {0.55,0.85,...,6.65}
        \draw[black!45, line width=.38pt] (\x,0.63) -- (\x+0.58,1.27);
\end{scope}
\begin{scope}
    \clip (1.05,0.75) rectangle (3.85,1.15);
    \foreach \x in {1.12,1.32,...,3.92}
        \foreach \y in {0.81,0.90,0.99,1.08}
            \fill[black!45] (\x,\y) circle[radius=.016];
\end{scope}
\foreach \x in {1.61,2.17,2.73,3.29,3.85,4.41,4.97,5.53,6.09}
    \draw[white, line width=.55pt] (\x,0.75) -- (\x,1.15);

\node[font=\tiny, text=black!65] at (2.45,0.47) {pre-2011};
\node[font=\tiny, text=black!65] at (5.25,0.47) {post-2011};
\draw[dashed, black!35] (3.85,0.63) -- (3.85,1.92);

\fill[synthOne] (0.95,0.09) rectangle (1.20,0.25);
\node[legendtext] at (1.28,0.17) {$\mathcal{T}_1$ in $A$};
\fill[synthOneLight] (3.35,0.09) rectangle (3.60,0.25);
\begin{scope}
    \clip (3.35,0.09) rectangle (3.60,0.25);
    \foreach \x in {3.25,3.37,...,3.60}
        \draw[black!45, line width=.3pt] (\x,0.05) -- (\x+0.18,0.31);
\end{scope}
\node[legendtext] at (3.68,0.17) {$\mathcal{T}_1$ in $B$};
\fill[synthTwo] (0.95,-0.19) rectangle (1.20,-0.03);
\foreach \x in {1.01,1.08,1.15}
    \foreach \y in {-0.15,-0.09,-0.05}
        \fill[white] (\x,\y) circle[radius=.011];
\node[legendtext] at (1.28,-0.11) {$\mathcal{T}_2$ in $A$};
\fill[synthTwoLight] (3.35,-0.19) rectangle (3.60,-0.03);
\begin{scope}
    \clip (3.35,-0.19) rectangle (3.60,-0.03);
    \foreach \x in {3.25,3.37,...,3.60}
        \draw[black!45, line width=.3pt] (\x,-0.23) -- (\x+0.18,0.03);
\end{scope}
\foreach \x in {3.41,3.48,3.55}
    \foreach \y in {-0.15,-0.09,-0.05}
        \fill[black!45] (\x,\y) circle[radius=.011];
\node[legendtext] at (3.68,-0.11) {$\mathcal{T}_2$ in $B$};

\end{tikzpicture}%
}
\caption{Corpus compositions for synthetic datasets.}
\label{fig:synthetic_corpus_composition}
\end{figure}

We evaluate the methods by recovering {\it known ground-truth hypotheses} from synthetic datasets.
We choose a dataset containing bill summaries from the 110-114th U.S. Congress~\citep{hoyle-etal-2022-neural}.
It has human-labeled high-level topics and granular subtopics, and metadata such as bill creation time, allowing us to repurpose it to construct synthetic corpora with \emph{controlled covariate structure}.
We instantiate the two data scenarios demonstrating the two challenges from \S\ref{sec:method}: \textbf{stratum imbalance} and \textbf{sign reversal}.
Figure~\ref{fig:synthetic_corpus_composition} summarizes their corpus compositions.

\subsubsection{Scenario 1: Stratum Imbalance}
\label{sec:exp_conditional}

We investigate whether methods can recover a \emph{targeted subset} of differences when many differences are simultaneously present.
We split the topics into two themes: \textit{Government \& Economy}, $\mathcal{T}_{gov}$ and \textit{Social policy}, $\mathcal{T}_{soc}$.
For each seed run, we randomly select three topics from $\mathcal{T}_{gov}$ and three topics from $\mathcal{T}_{soc}$.
For each topic $T_i$ (e.g., ``Health''), we randomly select one subtopic for $A$ and one subtopic for $B$ (e.g., ``Mental'' and ``Drug Industry'').
The example ground truth is in the form of ``is about Health: Mental''.
The targeted difference is in the social policy theme, and we thus evaluate recovery of ground truth under $\mathcal{T}_{soc}$.

To simulate the case when relevant strata are rare, an imbalance ratio $r$ controls the frequency of $\mathcal{T}_{soc}$ topics in $A$: 
each $\mathcal{T}_{soc}$ topic contributes at rate $r$ and each $\mathcal{T}_{gov}$ topic at rate $1{-}r$, with $B$ receiving the mirror allocation.
Lower $r$ suppresses the $\mathcal{T}_{soc}$ signal in $A$ more, making it harder to recover the targeted difference.
We pass three binary topic covariates to covariate-aware methods, one indicator per $\mathcal{T}_{soc}$ topic (more data details in Appendix~\ref{app:cond_data}).
These covariates should steer the methods toward the relevant strata, but do not reveal which subtopic distinguishes $A$ from $B$ within each topic.

\subsubsection{Scenario 2: Sign Reversal}
\label{sec:exp_confound}

We then investigate whether methods can recover targeted differences when a covariate masks the aggregate signal.
The covariate is the bill creation time period (binary: pre/post 2011).
For each seed run, we randomly select $10$ subtopics, each from a distinct topic, and split them into two groups $\mathcal{T}_1$ and $\mathcal{T}_2$ of five each.
In the pre-2011 period, $\mathcal{T}_1$ subtopics are assigned to $A$ and $\mathcal{T}_2$ to $B$; in the post-2011 period, the assignment reverses.
We pass one binary time-period covariate (pre-2011${=}1$) to covariate-aware methods,
and evaluate recovery of the five $\mathcal{T}_1$ subtopic differences.
By construction, the difference reverses sign across periods, violating NQI assumption.
We expect interaction-lasso to be the appropriate method, while demeaning-based methods are expected to fail.

\begin{table*}[t]
\centering
\small
\begin{tabular}{lrrrrrr}
\toprule
Method & \multicolumn{2}{c}{$r{=}0.50$} & \multicolumn{2}{c}{$r{=}0.35$} & \multicolumn{2}{c}{$r{=}0.20$} \\
\cmidrule(lr){2-3}\cmidrule(lr){4-5}\cmidrule(lr){6-7}
       & Surf. Sim. & F1 & Surf. Sim. & F1 & Surf. Sim. & F1 \\
\midrule
oracle
  & .750$\pm$.12 & .711$\pm$.08 & .737$\pm$.11 & .682$\pm$.06 & .723$\pm$.15 & .684$\pm$.05 \\
random
  & .280$\pm$.18 & .248$\pm$.10 & .400$\pm$.10 & .299$\pm$.09 & .347$\pm$.10 & .228$\pm$.08 \\
\midrule
\multicolumn{7}{l}{\textit{LLM-direct}} \\
llm-covariate
  & .330$\pm$.23 & .410$\pm$.11 & .137$\pm$.12 & .178$\pm$.11 & .057$\pm$.08 & .080$\pm$.03 \\
llm-global
  & .323$\pm$.16 & .325$\pm$.08 & .127$\pm$.11 & .155$\pm$.06 & .067$\pm$.09 & .090$\pm$.03 \\
\midrule
\multicolumn{7}{l}{\textit{SAE}} \\
lasso
  & .557$\pm$.18 & .394$\pm$.14 & .383$\pm$.13 & .235$\pm$.08 & .393$\pm$.12 & .230$\pm$.07 \\
separation score
  & .487$\pm$.17 & .340$\pm$.16 & .397$\pm$.16 & .221$\pm$.10 & .333$\pm$.13 & .228$\pm$.05 \\
\midrule
\multicolumn{7}{l}{\textit{Proposed}} \\
demeaned-reweighted-lasso
  & $^{\dagger\dagger}$\textbf{.700}$\pm$.11 & $^{\dagger\dagger}$\textbf{.533}$\pm$.12
  & $^{\dagger\dagger}$\textbf{.730}$\pm$.14 & $^{\dagger\dagger}$\textbf{.605}$\pm$.14
  & $^{\dagger\dagger}$\textbf{.730}$\pm$.11 & $^{\dagger\dagger}$\textbf{.534}$\pm$.10 \\
demeaned-lasso (ablation)
  & .530$\pm$.17 & .406$\pm$.16
  & $^{\dagger\dagger}$.533$\pm$.15 & $^{\dagger\dagger}$.429$\pm$.08
  & $^{\dagger}$.517$\pm$.13 & $^{\dagger\dagger}$.378$\pm$.07 \\
interaction-lasso
  & .407$\pm$.15 & .280$\pm$.09 & .363$\pm$.16 & .226$\pm$.12 & .357$\pm$.12 & .211$\pm$.10 \\
\bottomrule
\end{tabular}
\caption{Stratum Imbalance: $|\mathcal{H}|{=}3$.
  $\dagger$/$\dagger\dagger$: $p{<}0.05$/$0.01$ vs.\ lasso, exact paired sign-flip test.}
\label{tab:scenario1}
\end{table*}

\subsection{Evaluation Metrics}
\label{sec:evaluation}
We evaluate how well the methods recover the targeted ground-truth differences on the held-out set.
For each scenario, we use an \llm to annotate generated hypotheses on held-out texts and match them to reference hypotheses using the Hungarian algorithm on reference-vs.-generated annotation correlations.
See Appendix~\ref{app:evaluation_metrics} for detailed process.
We report two metrics computed over matched pairs. 
\textbf{Surface similarity}: for each matched reference/inferred hypothesis pair, we prompt \texttt{gpt-4.1} to assess whether the two hypotheses are the \emph{same}, \emph{related}, or \emph{distinct}, with corresponding scores of 1.0, 0.5, and 0.0 (prompt in Table~\ref{tab:surface_similarity_prompt}, examples in Table~\ref{tab:surface_similarity_examples}).
To improve stability, we sample five outputs at temperature $0.7$ and average them, then report the mean score across $|\mathcal{H}|$ pairs.
\textbf{F1 similarity}: for each matched pair, we compute the F1-score between the reference and inferred hypothesis {\it text-level} annotations on the held-out set.

\subsection{Baselines}
\label{sec:baselines}
Since no existing method handles covariates, we compare against SAE-based methods~\citep{movva2025sparse} as the primary baselines. 
For fair comparison, we use \texttt{gpt-4.1} for proposing hypotheses if not specified otherwise, and OpenAI’s text-embedding-3-small for embeddings.

\noindent\textbf{\sae-based.} (1) \textbf{\lasso}: use Eq.~\ref{eq:lasso} to select the top $|\mathcal{H}|$ features.
(2) \textbf{separation score}: select features by the mean of $y$ among the highest-activating texts vs. zero-activating texts.\footnote{https://github.com/rmovva/HypotheSAEs}

\noindent\textbf{\llm-direct.} We include direct prompting baselines to test whether covariate-aware discovery can be obtained simply by showing sampled texts to an \llm, without a corpus-level feature-selection step.
\textbf{llm-global} is prompted with sampled corpus texts, while \textbf{llm-covariate} is also shown covariate-profile labels for each text.
Because these baselines only observe the prompt sample, we use a generous sample budget in the main tables.
Appendix~\ref{app:llm_direct_sensitivity} gives the sensitivity analysis of the sample size.

\noindent\textbf{Reference baselines.} To calibrate the evaluation, we include:
\textbf{Oracle} as an upper reference that gives the \llm $20$ examples sampled from the exact positive and negative sides of each ground-truth contrast.
\textbf{Random} samples $20$ training texts into arbitrary positive and negative halves before the same interpretation step, measuring how much recovery can come from generic topical cues alone.
We use $20$ examples to match the SAE interpretation stage.
More implementation details for these baseline methods are provided in Appendix~\ref{app:method_implementation_details}.

\noindent\textbf{Proposed methods.} We compare \textbf{interaction-lasso} (\S\ref{sec:interaction_lasso}),
\textbf{demeaned-reweighted-lasso} (\S\ref{sec:demeaned_lasso}), and
\textbf{demeaned-lasso} as ablation with the baselines.

\noindent\textbf{Statistical testing.}
Each proposed method is compared to \textbf{\lasso} on seed-level surface similarity and F1, with $10$ seeds per setting.
We test whether the mean paired difference differs from zero with an exact two-sided sign-flip test.

\begin{table}[t]
  \centering
  \small
  \begin{tabular}{lrr}
  \toprule
  Method & Surf. Sim. & F1 \\
  \midrule
  oracle & .716$\pm$.10 & .788$\pm$.06 \\
  random & .418$\pm$.12 & .381$\pm$.10 \\
  \midrule
  \multicolumn{3}{l}{\textit{LLM-direct}} \\
  llm-global            & .210$\pm$.10 & .269$\pm$.05 \\
  llm-covariate         & .224$\pm$.09 & .325$\pm$.16 \\
  \midrule
  \multicolumn{3}{l}{\textit{SAE}} \\
  lasso                 & .484$\pm$.06 & .362$\pm$.09 \\
  sep.\ score           & .460$\pm$.11 & .379$\pm$.12 \\
  \midrule
  \multicolumn{3}{l}{\textit{Proposed}} \\
  interaction-lasso
    & $^{\dagger\dagger}$\textbf{.600}$\pm$.10 & $^{\dagger\dagger}$\textbf{.576}$\pm$.08 \\
  demeaned-reweighted-lasso & .448$\pm$.14 & .335$\pm$.11 \\
  demeaned-lasso (ablation) & .454$\pm$.11 & .398$\pm$.12 \\
  \bottomrule
  \end{tabular}
  \caption{Sign reversal: $|\mathcal{H}|{=}5$.
    $\dagger\dagger$: $p{<}0.01$ vs.\ lasso, exact paired sign-flip test.}
  \label{tab:scenario2}
  \end{table}

\subsection{Results and Discussion}
\label{sec:syn_results}

\paragraph{Demeaned-reweighted-lasso steers discovery toward specified covariates and recovers suppressed within-stratum signals.}
Table~\ref{tab:scenario1} evaluates conditional discovery as the target social-policy strata become increasingly underrepresented.
Across all imbalance levels, demeaned-reweighted-lasso remains near the oracle in surface similarity and significantly improves F1 over \lasso, the strongest global \sae baseline.
Even at $r{=}0.50$, it reaches $.700$ surface similarity, near the oracle's $.750$ and above \lasso's $.557$ ($p{=}0.016$), showing that covariates steer feature selection toward the target $\mathcal{T}_{soc}$ differences.
The gap widens under stronger imbalance: \lasso falls to $.383$--$.393$ surface similarity, while demeaned-reweighted-lasso remains at $.730$.
The demeaned-lasso ablation improves over \lasso under imbalance but remains below the reweighted variant, indicating that residualizing within strata helps but reweighting further helps rare target strata from being suppressed.

\noindent\textbf{The covariate structure determines the appropriate selection model.}
Table~\ref{tab:scenario2} evaluates sign reversal, where the time-period covariate reverses the differences across strata and makes the aggregate signal misleading.
As \S\ref{sec:method} notes, residualization is appropriate when the covariate induces fixed stratum-level shifts, but not when the difference itself changes sign.
As expected, neither demeaned variant significantly improves over \lasso.

\noindent\textbf{Interaction-lasso accounts for sign reversal.}
Interaction-lasso is the only covariate-aware method that significantly improves over \lasso in Table~\ref{tab:scenario2}, increasing surface similarity from $.484$ to $.600$ ($p{=}0.008$) and F1 from $.362$ to $.576$ ($p{=}0.002$).
This suggests modeling feature-by-time interactions allows a feature to be selected when it is discriminative within a period even if its global effect cancels.

\noindent\textbf{Interaction-lasso faces computational and sparsity limitations.}
Despite this advantage in the sign-reversal setting, interaction-lasso does not improve over global SAE baselines in Table~\ref{tab:scenario1}, despite being the more general model.
This empirically validates our discussion in \S\ref{sec:interaction_lasso} that interaction-lasso faces feature-space expansion and sparsity limitations.
These limitations may also help explain why interaction-lasso is best in Table~\ref{tab:scenario2} but still falls short of the oracle.

\noindent\textbf{Statistical modeling of covariates, not prompt-level exposure, drives gains.}
Both tables show that \llm-direct methods fall behind \sae-based methods.
The sensitivity analysis in Appendix~\ref{app:llm_direct_sensitivity} further shows that llm-covariate does not consistently outperform llm-global across sample sizes or with a stronger reasoning model (\texttt{gpt-5.4}).

\section{Validation on Real-World Datasets}
\label{sec:realworld}

The synthetic experiments above test recovery under known conditional structure.
For real-world corpora, where no ground-truth hypotheses exist, we evaluate whether conditional generation surfaces hypotheses that domain experts find useful for comparing outcome groups within researcher-specified covariate strata.
Because our goal is to generate hypotheses discriminative within strata of $\mathbf{C}$ while retaining globally discriminative findings, we focus on hypotheses unique to the covariate-aware method relative to the global baseline.

\subsection{Real-World Datasets}
We include two real-world datasets and instantiate $\mathbf{C}$ with one researcher-specified binary covariate for each, chosen from domain knowledge:
(1) \congress~\citep{gentzkow2010drives}: U.S. speeches from the 109th Congress (2005-07), a longstanding application in computational social science~\citep{grimmer2021machine}.
The outcome variable $y$ is party affiliation (Republican vs. Democrat).
The covariate indicates whether a speech contains substantive public-policy discussion rather than procedural talk (e.g., scheduling and unanimous-consent requests).
Congressional speeches mix these forms of activities, while procedural talk does not reflect the substantive policy disagreements that political scientists typically seek to measure~\citep{gentzkow2019measuring}.
Conditioning on this covariate allows us to guide hypothesis generation to reflect policy discussion beyond procedural activity.
(2) \ncte~\citep{demszky2023ncte}: the largest publicly available dataset of math classroom transcripts with rich expert annotations of instructional quality~\cite{hill2008mathematical} and classroom environment~\cite{pianta2012classroom},
collected by the National Center for Teacher Effectiveness (\ncte) 
between 2010-2013.
The outcome variable $y$ is low vs. high quality \remed, a measure of how well teachers remediate student math errors and difficulties in class.
This teaching practice is central to math instruction as it provides students with opportunities for conceptual learning~\cite{bray2011collective}.
We use behavioral management (e.g., organizing activities and redirecting students) as the covariate because management and instruction are intertwined: teachers often need to minimize time spent on behavioral management to focus on content instruction~\cite{emmer2003classroom}.
Conditioning on this covariate allows us to examine whether the model surfaces remediation-oriented 
teaching practices beyond behavioral management.

\subsection{Study Setup}
We expect the two covariates to steer hypothesis discovery toward within-stratum differences, analogous to the stratum-imbalance setting in \S\ref{sec:exp_conditional}.
Prior work suggests that sign reversal is rare in practice~\citep{cochrane2024handbook}, so we compare the global \lasso baseline with demeaned-reweighted-lasso as the covariate-aware method.
Our main metric is the \emph{helpfulness} of each hypothesis, on a 1--5 scale, for understanding the group difference while accounting for the specified covariate.
Each hypothesis is accompanied by its prevalence in the two outcome groups.
After rating helpfulness, annotators reviewed the same hypotheses with global and covariate-stratified statistics and rated \emph{conditional interpretive value} on a 1--5 scale: whether the breakdown by the specified covariate added interpretive value beyond the global finding.
We set $|\mathcal{H}|{=}10$ to limit the cognitive burden on annotators.

We recruit research scholars for each domain. 
Specifically, we invited two computational social science scholars familiar with U.S. politics to annotate \congress. 
And we invited two educational researchers who have used \ncte in prior research to annotate the \ncte dataset. 
More details on annotation instruction, expert recruitment, and covariate operationalization are in Appendix~\ref{app:real_world_validation}.

\subsection{Results and Discussion}

To focus on complementary discoveries, we manually match semantically similar hypotheses across the two method outputs.
Matched hypotheses capture patterns recovered by both methods and are therefore less diagnostic of the added value of covariate-aware selection.
We report the full lists and matching decisions for both datasets in Appendix~\ref{app:user_study_results}.
Differences among matched items may also reflect LLM interpretation noise, such as wording specificity.
Table~\ref{tab:ncte_unique_hypotheses} lists the unique \ncte hypotheses by method as qualitative examples; aggregate expert ratings appear separately in Table~\ref{tab:real_unique_ratings}.

\noindent\textbf{Covariate-aware selection surfaces useful within-stratum hypotheses.}
Table~\ref{tab:ncte_unique_hypotheses} illustrates this shift in \ncte: unique covariate-aware hypotheses move from classroom-management context toward remediation or other instructional activities.
The global \lasso baseline surfaced two unique hypotheses about classroom management, the covariate-defined context we aim to account for.
By contrast, the covariate-aware method surfaced more instruction-oriented hypotheses about units-of-weight discussion, individualized follow-up, and peer explanation.
The small-group hypothesis is especially illustrative: rather than only identifying group organization, it emphasizes teachers prompting students to help and explain concepts to one another.
The same pattern appears more modestly in \congress: among the unique hypotheses, the global baseline retains a procedural hypothesis, while the covariate-aware method surfaces policy-related hypotheses about economic performance and border security.
Together, these examples suggest that covariate-aware selection can surface unique hypotheses aligned with the researcher-specified strata.
The averaged ratings in Table~\ref{tab:real_unique_ratings} further confirm that these hypotheses are more useful than their counterparts ($3.10$ vs. $2.50$).

{
\renewcommand{\arraystretch}{1.3} %

\begin{table}[t]
  \centering
  \footnotesize
  \begin{tabular}{p{0.97\linewidth}}
  \toprule
  Unique \lasso hypotheses \\
  \midrule
  {\it direct students for logistical tasks such as collecting items.} \\
  {\it prompt student to write reflections about learning at the end.} \\
  {\it use management language to enforce compliance.} \\
  \midrule
  Unique demeaned-reweighted-lasso hypotheses \\
  \midrule
  {\it Extended discussions of units of weight.} \\
  {\it follow-up with specific students about needs.} \\
  {\it organize to work in small groups and prompt students help and explain concepts to one another.} \\
  \bottomrule
  \end{tabular}
  \caption{NCTE hypotheses unique to each method after semantic matching.
  Hypotheses are shortened for readability. Full wording appears in Appendix~\ref{app:user_study_results}.}
  \label{tab:ncte_unique_hypotheses}
\end{table}
}

\begin{table}[t]
  \centering
  \small
  \begin{tabular}{lcc}
  \toprule
  Unique hypotheses & Helpfulness & Cond.\\
  \midrule
  \multicolumn{3}{l}{\textit{Congress}} \\
  \lasso & 2.75 & 2.50 \\
  demeaned-reweighted-lasso & 3.25 & 1.50 \\
  \midrule
  \multicolumn{3}{l}{\textit{NCTE}} \\
  \lasso & 2.33 & 2.33 \\
  demeaned-reweighted-lasso & 3.00 & 3.50 \\
  \midrule
  \multicolumn{3}{l}{\textit{Overall}} \\
  \lasso & 2.50 & 2.40 \\
  demeaned-reweighted-lasso & 3.10 & 2.70 \\
  \bottomrule
  \end{tabular}
  \caption{Expert ratings for hypotheses unique to each method after semantic matching.
  Helpfulness and conditional interpretive value (Cond.) are 1--5 Likert means.}
  \label{tab:real_unique_ratings}
\end{table}

\paragraph{The value of the covariate depends on the dataset.}
Conditional interpretive value measures whether the $\mathbf{C}$-stratified statistics change how annotators interpret a generated hypothesis.
In \ncte, behavioral management helps distinguish classroom organization from instructional activities, and unique covariate-aware hypotheses received higher conditional-value ratings than unique \lasso hypotheses (3.50 vs. 2.33).
In \congress, the policy/procedure covariate was less informative as a displayed breakdown: procedural talk remained strong in both method outputs,
and the two unique covariate-aware hypotheses were already clearly policy-related from their text.
As one annotator noted, the policy/procedure breakdown often confirmed what was already expected rather than changing the interpretation.
Accordingly, unique covariate-aware hypotheses in \congress were rated as more helpful (3.25 vs.\ 2.75) but lower in conditional interpretive value (1.50 vs.\ 2.50).

\section{Related Work}
\label{sec:related}

\paragraph{LLM-based hypothesis discovery.}
Complementing the task overview in \S\ref{sec:background}, we situate \llm-based hypothesis generation among classical text-analysis tools and applications.
Classical text analysis methods such as n-gram frequency comparisons
\citep{gentzkow2019measuring} and topic models, including LDA~\citep{blei2003latent} and structural topic models~\citep{roberts2014structural}, remain popular tools
for relating text to target variables~\citep{grimmer2022text}, but typically
produce word lists that require further interpretation.
Recent \llm-based methods instead generate natural-language descriptions of
corpus differences from sampled examples~\citep{zhong2022describing,zhong2023goal}, iterative error analysis~\citep{zhou2024hypothesis},
literature-grounded candidates~\citep{liu-etal-2025-literature}, or statistical modeling~\citep{zhong2024nlparam,movva2025sparse,agarwal2026autodiscovery}.
Related work has applied this idea to settings such as headline click rates~\citep{batista2024words}, social media engagement~\citep{xu2026does}, and causal measurement from text~\citep{modarressi2025causal}.
OpenAI recently released a text measurement toolkit that also supports hypothesis discovery~\citep{asirvatham2026gpt}, demonstrating its potential for real-world applications.
Our work differs by incorporating domain knowledge through statistical models of covariate structure, steering discovery toward researcher-specified differences of interest.

\paragraph{Statistical methods for covariate adjustment.}
Our methods draw on standard statistical and econometric tools for making comparisons conditional on covariates. 
Fixed effects and within-group residualization are classical ways to remove nuisance variation~\citep{angrist2009mostly}, with the Frisch--Waugh--Lovell theorem providing the corresponding partialling-out justification~\citep{lovell1963seasonal}. 
Simpson's paradox~\citep{simpson1951interpretation} shows why such adjustment matters:
marginal associations can reverse relative to associations within each stratum~\citep{blyth1972simpson}.
When conditional effects change sign across strata, the problem becomes one of
qualitative interaction~\citep{gail1985testing}. These results provide the
statistical basis for using residualization, reweighting, and interaction terms
as principled feature-selection mechanisms for conditional hypothesis
generation.

\section{Conclusion}
\label{sec:conclusion}

We introduced conditional hypothesis generation for text analysis, a framework for steering natural-language hypothesis discovery toward differences that hold within researcher-specified covariate strata.
This setting makes explicit two statistical challenges: \emph{stratum imbalance}, where signals from underrepresented strata are suppressed, and \emph{sign reversal}, where aggregation can cancel or misrepresent within-stratum differences.
By incorporating covariates into the \sae feature selection, our methods let researchers encode domain knowledge about which strata should shape discovery.

We proposed two complementary covariate-aware methods.
Interaction-lasso models feature--covariate interactions and is suited to sign reversal, while demeaned-reweighted-lasso removes stratum-level variation and reweights underrepresented strata to recover within-stratum signals under a consistent-direction assumption.
Synthetic experiments show that these modeling choices matter: demeaned-reweighted-lasso recovers hypotheses suppressed by stratum imbalance, whereas interaction-lasso is needed when the direction of the difference reverses.
Expert validation on two real-world datasets, \congress and \ncte, further suggests that covariate-aware method can surface more useful hypotheses for researchers' stated comparisons than those unique to a global baseline.

\clearpage
\section*{Limitations}

\paragraph{Dependence on researcher-specified covariates.}
Our framework intentionally makes covariate specification a design choice: researchers decide which contexts should shape hypothesis discovery based on domain knowledge and the comparison they want to make.
This choice gives researchers control over the target of discovery, but it also means the methods are only as useful as the covariates supplied.
They can reduce the influence of known covariate structure, but they cannot discover which covariates should matter, correct for omitted covariates, or make the resulting hypotheses causal claims.
Poorly chosen or noisy covariates may steer discovery toward unhelpful strata, and overly fine-grained covariates may create sparse cells where within-stratum comparisons are unreliable.

\paragraph{Categorical strata.}
Both proposed methods are designed around observed covariate strata.
In this paper we focus on binary or categorical covariates, such as policy/procedure status, behavior-management level, topic indicators, and time period.
Continuous covariates would require discretization or a different residualization strategy,
The quality of the conditional hypotheses therefore depends on the validity of the covariate operationalization, including any automated annotations used to define covariates.

\paragraph{Method-specific assumptions.}
The two methods address different statistical regimes.
Demeaned-reweighted-lasso assumes no qualitative interaction: the relevant group difference should have a consistent direction across covariate strata.
When the direction reverses, as in our sign-reversal experiment, residualization can remove nuisance variation but cannot recover the stratum-specific effects of interest.
Interaction-lasso is better suited to such cases, but it expands the feature space and can be unstable when both \sae activations and covariate strata are sparse.
In practice, researchers should inspect stratum-level statistics when possible and choose the selection model based on whether sign reversal is plausible.

\paragraph{Dependence on the \sae interpretation pipeline.}
Our implementation builds on the \sae-based hypothesis-generation pipeline of~\citet{movva2025sparse}.
It therefore inherits the pipeline's requirements and failure modes~\cite{peng2025use}: access to the full corpus for feature extraction, a trained \sae whose features are sufficiently interpretable for the domain, hyperparameter choices that affect the granularity of learned features, and an \llm interpretation step that can phrase selected features imperfectly.
Our contribution modifies feature selection, not the underlying representation learning or natural-language verbalization stages.

\paragraph{Evaluation of discovery quality.}
The synthetic experiments provide controlled ground truth for stratum imbalance and sign reversal, but their reference hypotheses are constructed from bill-summary topics and may not cover the full range of linguistic phenomena or covariate choices encountered in applied text analysis.
The real-world validation complements these experiments with expert judgments in two substantively different domains, \congress and \ncte, while broader generalization will require additional datasets, covariates, and expert panels.
Such validation is difficult because scientific discovery has no single objective metric: what counts as a useful hypothesis depends on domain expertise, the research question, and the implicit norms of the relevant research community rather than explicit deductive criteria alone~\citep{polanyi2000republic}.
The ratings should therefore be interpreted as evidence that covariate-aware selection can surface useful hypotheses, not as a complete benchmark of performance across domains.
Future evaluations should test more covariates, multi-valued and continuous settings, larger expert panels, and downstream research workflows where generated hypotheses are refined and validated by domain specialists.

\bibliography{custom}

@article{agarwal2026autodiscovery,
  title={Autodiscovery: Open-ended scientific discovery via bayesian surprise},
  author={Agarwal, Dhruv and Majumder, Bodhisattwa Prasad and Adamson, Reece and Chakravorty, Megha and Gavireddy, Satvika Reddy and Parashar, Aditya and Surana, Harshit and Dalvi Mishra, Bhavana and McCallum, Andrew and Sabharwal, Ashish and others},
  journal={Advances in Neural Information Processing Systems},
  volume={38},
  pages={25181--25219},
  year={2026}
}

@article{peng2025use,
  title={Use sparse autoencoders to discover unknown concepts, not to act on known concepts},
  author={Peng, Kenny and Movva, Rajiv and Kleinberg, Jon and Pierson, Emma and Garg, Nikhil},
  journal={arXiv preprint arXiv:2506.23845},
  year={2025}
}

@article{kane2012gathering,
  title={Gathering Feedback for Teaching: Combining High-Quality Observations with Student Surveys and Achievement Gains. Research Paper. MET Project.},
  author={Kane, Thomas J and Staiger, Douglas O},
  journal={Bill \& Melinda Gates Foundation},
  year={2012},
  publisher={ERIC}
}

@inproceedings{xu2024promises,
  title={The promises and pitfalls of using language models to measure instruction quality in education},
  author={Xu, Paiheng and Liu, Jing and Jones, Nathan and Cohen, Julie and Ai, Wei},
  booktitle={Proceedings of the 2024 Conference of the North American Chapter of the Association for Computational Linguistics: Human Language Technologies (Volume 1: Long Papers)},
  pages={4375--4389},
  year={2024}
}

@inproceedings{xu2026does,
  title={Does Geo-co-location Matter? A Case Study of Public Health Conversations during COVID-19},
  author={Xu, Paiheng and Raschid, Louiqa and Frias-Martinez, Vanessa},
  booktitle={Proceedings of the International AAAI Conference on Web and Social Media},
  volume={20},
  number={1},
  pages={2518--2537},
  year={2026}
}

@article{polanyi2000republic,
  title={The republic of science: its political and economic theory Minerva, I (1)(1962), 54-73},
  author={Polanyi, Michael and Ziman, John and Fuller, Steve},
  journal={Minerva},
  volume={38},
  number={1},
  pages={1--32},
  year={2000},
  publisher={JSTOR}
}

@article{crump2009dealing,
  title={Dealing with limited overlap in estimation of average treatment effects},
  author={Crump, Richard K and Hotz, V Joseph and Imbens, Guido W and Mitnik, Oscar A},
  journal={Biometrika},
  volume={96},
  number={1},
  pages={187--199},
  year={2009},
  publisher={Oxford University Press}
}

@article{bien2013lasso,
  title={A lasso for hierarchical interactions},
  author={Bien, Jacob and Taylor, Jonathan and Tibshirani, Robert},
  journal={Annals of statistics},
  volume={41},
  number={3},
  pages={1111},
  year={2013}
}

@article{hill2008mathematical,
  title={Mathematical knowledge for teaching and the mathematical quality of instruction: An exploratory study},
  author={Hill, Heather C and Blunk, Merrie L and Charalambous, Charalambos Y and Lewis, Jennifer M and Phelps, Geoffrey C and Sleep, Laurie and Ball, Deborah Loewenberg},
  journal={Cognition and instruction},
  volume={26},
  number={4},
  pages={430--511},
  year={2008},
  publisher={Taylor \& Francis}
}

@incollection{emmer2003classroom,
  title={Classroom management: A critical part of educational psychology, with implications for teacher education},
  author={Emmer, Edmund T and Stough, Laura M},
  booktitle={Educational psychology},
  pages={103--112},
  year={2003},
  publisher={Routledge}
}

@article{bray2011collective,
  title={A collective case study of the influence of teachers' beliefs and knowledge on error-handling practices during class discussion of mathematics},
  author={Bray, Wendy S},
  journal={Journal for Research in Mathematics education},
  volume={42},
  number={1},
  pages={2--38},
  year={2011},
  publisher={National Council of Teachers of Mathematics}
}

@book{pianta2012classroom,
  title={Classroom Assessment Scoring System: CLASS: Upper Elementary Manual},
  author={Pianta, Robert C and Hamre, Bridget K and Mintz, Susan},
  year={2012},
  publisher={Teachstone}
}

@inproceedings{demszky2023ncte,
  title={The NCTE transcripts: A dataset of elementary math classroom transcripts},
  author={Demszky, Dorottya and Hill, Heather},
  booktitle={Proceedings of the 18th Workshop on Innovative Use of NLP for Building Educational Applications (BEA 2023)},
  pages={528--538},
  year={2023}
}

@article{grimmer2021machine,
  title={Machine learning for social science: An agnostic approach},
  author={Grimmer, Justin and Roberts, Margaret E and Stewart, Brandon M},
  journal={Annual Review of Political Science},
  volume={24},
  number={1},
  pages={395--419},
  year={2021},
  publisher={Annual Reviews}
}

@article{gentzkow2010drives,
  title={What drives media slant? Evidence from US daily newspapers},
  author={Gentzkow, Matthew and Shapiro, Jesse M},
  journal={Econometrica},
  volume={78},
  number={1},
  pages={35--71},
  year={2010},
  publisher={Wiley Online Library}
}

@misc{adler2018congressional,
  title={Congressional Bills Project: 1995-2018},
  author={Adler, E Scott and Wilkerson, John},
  year={2018},
  publisher={NSF}
}

@inproceedings{hoyle-etal-2022-neural,
    title = "Are Neural Topic Models Broken?",
    author = "Hoyle, Alexander  and
      Goel, Pranav  and
      Sarkar, Rupak  and
      Resnik, Philip",
    editor = "Goldberg, Yoav  and
      Kozareva, Zornitsa  and
      Zhang, Yue",
    booktitle = "Findings of the Association for Computational Linguistics: EMNLP 2022",
    month = dec,
    year = "2022",
    address = "Abu Dhabi, United Arab Emirates",
    publisher = "Association for Computational Linguistics",
    url = "https://aclanthology.org/2022.findings-emnlp.390/",
    doi = "10.18653/v1/2022.findings-emnlp.390",
    pages = "5321--5344"
}

@article{modarressi2025causal,
  title={Causal inference on outcomes learned from text},
  author={Modarressi, Iman and Spiess, Jann and Venugopal, Amar},
  journal={arXiv preprint arXiv:2503.00725},
  year={2025}
}

@article{batista2024words,
  title={Words that work: Using language to generate hypotheses},
  author={Batista, Rafael M and Ross, James},
  journal={Available at SSRN 4926398},
  year={2024}
}

@article{blei2003latent,
  title={Latent dirichlet allocation},
  author={Blei, David M and Ng, Andrew Y and Jordan, Michael I},
  journal={Journal of machine Learning research},
  volume={3},
  number={Jan},
  pages={993--1022},
  year={2003}
}

@article{gentzkow2019measuring,
  title={Measuring group differences in high-dimensional choices: method and application to congressional speech},
  author={Gentzkow, Matthew and Shapiro, Jesse M and Taddy, Matt},
  journal={Econometrica},
  volume={87},
  number={4},
  pages={1307--1340},
  year={2019},
  publisher={Wiley Online Library}
}

@article{taddy2013multinomial,
  title={Multinomial inverse regression for text analysis},
  author={Taddy, Matt},
  journal={Journal of the American Statistical Association},
  volume={108},
  number={503},
  pages={755--770},
  year={2013},
  publisher={Taylor \& Francis}
}

@inproceedings{liu-etal-2025-literature,
    title = "Literature Meets Data: A Synergistic Approach to Hypothesis Generation",
    author = "Liu, Haokun  and
      Zhou, Yangqiaoyu  and
      Li, Mingxuan  and
      Yuan, Chenfei  and
      Tan, Chenhao",
    editor = "Che, Wanxiang  and
      Nabende, Joyce  and
      Shutova, Ekaterina  and
      Pilehvar, Mohammad Taher",
    booktitle = "Proceedings of the 63rd Annual Meeting of the Association for Computational Linguistics (Volume 1: Long Papers)",
    month = jul,
    year = "2025",
    address = "Vienna, Austria",
    publisher = "Association for Computational Linguistics",
    url = "https://aclanthology.org/2025.acl-long.12/",
    doi = "10.18653/v1/2025.acl-long.12",
    pages = "245--281",
    ISBN = "979-8-89176-251-0"
}

@inproceedings{zhou2024hypothesis,
  title={Hypothesis Generation with Large Language Models},
  author={Zhou, Yangqiaoyu and Liu, Haokun and Srivastava, Tejes and Mei, Hongyuan and Tan, Chenhao},
  booktitle={Proceedings of the 1st Workshop on NLP for Science (NLP4Science)},
  pages={117--139},
  year={2024}
}

@article{roberts2014structural,
  title={Structural topic models for open-ended survey responses},
  author={Roberts, Margaret E and Stewart, Brandon M and Tingley, Dustin and Lucas, Christopher and Leder-Luis, Jetson and Gadarian, Shana Kushner and Albertson, Bethany and Rand, David G},
  journal={American journal of political science},
  volume={58},
  number={4},
  pages={1064--1082},
  year={2014},
  publisher={Wiley Online Library}
}

@book{grimmer2022text,
  title={Text as data: A new framework for machine learning and the social sciences},
  author={Grimmer, Justin and Roberts, Margaret E and Stewart, Brandon M},
  year={2022},
  publisher={Princeton University Press}
}

@article{grimmer2013text,
  title={Text as data: The promise and pitfalls of automatic content analysis methods for political texts},
  author={Grimmer, Justin and Stewart, Brandon M},
  journal={Political analysis},
  volume={21},
  number={3},
  pages={267--297},
  year={2013},
  publisher={Cambridge University Press}
}

@phdthesis{card2019accelerating,
  title={Accelerating Text-as-Data Research in Computational Social Science},
  author={Card, Dallas},
  year={2019},
  school={University of Washington}
}

@article{simpson1951interpretation,
  title={The interpretation of interaction in contingency tables},
  author={Simpson, Edward H},
  journal={Journal of the Royal Statistical Society: Series B (Methodological)},
  volume={13},
  number={2},
  pages={238--241},
  year={1951},
  publisher={Wiley}
}

@InProceedings{zhong2022describing,
  title = 	 {Describing Differences between Text Distributions with Natural Language},
  author =       {Zhong, Ruiqi and Snell, Charlie and Klein, Dan and Steinhardt, Jacob},
  booktitle = 	 {Proceedings of the 39th International Conference on Machine Learning},
  pages = 	 {27099--27116},
  year = 	 {2022},
  editor = 	 {Chaudhuri, Kamalika and Jegelka, Stefanie and Song, Le and Szepesvari, Csaba and Niu, Gang and Sabato, Sivan},
  volume = 	 {162},
  series = 	 {Proceedings of Machine Learning Research},
  month = 	 {17--23 Jul},
  publisher =    {PMLR},
  url = 	 {https://proceedings.mlr.press/v162/zhong22a.html}
}

@article{zhong2023goal,
  title={Goal driven discovery of distributional differences via language descriptions},
  author={Zhong, Ruiqi and Zhang, Peter and Li, Steve and Ahn, Jinwoo and Klein, Dan and Steinhardt, Jacob},
  journal={Advances in Neural Information Processing Systems},
  volume={36},
  pages={40204--40237},
  year={2023}
}

@inproceedings{movva2025sparse,
  title={Sparse Autoencoders for Hypothesis Generation},
  author={Movva, Rajiv and Peng, Kenny and Garg, Nikhil and Kleinberg, Jon and Pierson, Emma},
  booktitle={International Conference on Machine Learning},
  pages={44997--45023},
  year={2025},
  organization={PMLR}
}

@article{zhong2024nlparam,
  title={Explaining datasets in words: Statistical models with natural language parameters},
  author={Zhong, Ruiqi and Wang, Heng and Klein, Dan and Steinhardt, Jacob},
  journal={Advances in Neural Information Processing Systems},
  volume={37},
  pages={79350--79380},
  year={2024}
}

@article{lovell1963seasonal,
  title={Seasonal adjustment of economic time series and multiple regression analysis},
  author={Lovell, Michael C},
  journal={Journal of the American Statistical Association},
  volume={58},
  number={304},
  pages={993--1010},
  year={1963},
  publisher={Taylor \& Francis}
}

@inproceedings{cunningham2023sparse,
  title={Sparse autoencoders find highly interpretable features in language models},
  author={Huben, Robert and Cunningham, Hoagy and Smith, Logan and Ewart, Aidan and Sharkey, Lee},
  booktitle={International Conference on Learning Representations},
  volume={2024},
  pages={7827--7845},
  year={2024}
}

@book{angrist2009mostly,
  title={Mostly harmless econometrics: An empiricist's companion},
  author={Angrist, Joshua D and Pischke, J{\"o}rn-Steffen},
  year={2009},
  publisher={Princeton university press}
}

@article{blyth1972simpson,
  title={On Simpson's paradox and the sure-thing principle},
  author={Blyth, Colin R},
  journal={Journal of the American Statistical Association},
  volume={67},
  number={338},
  pages={364--366},
  year={1972},
  publisher={Taylor \& Francis}
}

@article{gail1985testing,
  title={Testing for qualitative interactions between treatment effects and patient subsets},
  author={Gail, Mitchell and Simon, Richard},
  journal={Biometrics},
  pages={361--372},
  year={1985},
  publisher={JSTOR}
}

@book{cochrane2024handbook,
  title={Cochrane Handbook for Systematic Reviews of Interventions},
  author={Higgins, Julian P. T. and Thomas, James and Chandler, Jacqueline and Cumpston, Miranda and Li, Tianjing and Page, Matthew J. and Welch, Vivian A.},
  edition={Version 6.5},
  year={2024},
  publisher={Cochrane},
  url={https://training.cochrane.org/handbook/current}
}

@techreport{asirvatham2026gpt,
  title={{GPT} as a Measurement Tool},
  author={Asirvatham, Hemanth and Mokski, Elliott and Shleifer, Andrei},
  institution={National Bureau of Economic Research},
  type={Working Paper},
  number={34834},
  year={2026},
  doi={10.3386/w34834}
}

\appendix
\section{Synthetic Dataset Details}
\label{app:cond_data}

Topics are drawn from the BILLS dataset~\citep{adler2018congressional,hoyle-etal-2022-neural}, which provides 21 high-level topics and 114 low-level subtopics.
For Scenario 1: conditional discovery in Section~\ref{sec:exp_conditional}, we define two thematic pools: $\mathcal{T}_{gov}$ (Government \& Economy) and $\mathcal{T}_{soc}$ (Social Policy).
A topic is eligible if it has at least two subtopics each with $\geq 320$ samples, ensuring sufficient data to support the imbalanced sampling under the largest imbalance ratio tested.
The eligible topics retained for each theme are listed in Table~\ref{tab:eligible_topics}.
Each Scenario 1 corpus contains $N{=}2{,}400$ bill summaries, balanced across $A$ and $B$ ($N_A{=}N_B{=}1{,}200$).
For each selected topic, $400$ examples are allocated across the two corpora.
For example, when $r{=}0.20$, corpus $A$ contains $3{\times}400{\times}0.20{=}240$ examples from $\mathcal{T}_{soc}$ and $3{\times}400{\times}0.80{=}960$ examples from $\mathcal{T}_{gov}$, with corpus $B$ receiving the mirror allocation.
Each Scenario 2 corpus contains $N{=}3{,}000$ bill summaries ($N_A{=}N_B{=}1{,}500$), with equal-sized pre-2011 and post-2011 strata so that the within-period signals cancel at the aggregate level.

\begin{table}[h]
\centering
\footnotesize
\begin{tabular}{ll}
\toprule
\textbf{Theme} & \textbf{Eligible Topics} \\
\midrule
$\mathcal{T}_{gov}$ & Government Operations, Defense, \\
(Gov.\ \& Econ.) & Domestic Commerce, Macroeconomics, \\
 & Law and Crime, Energy, Transportation \\
\midrule
$\mathcal{T}_{soc}$ & Health, Education, \\
(Social Policy) & Environment, Public Lands \\
\bottomrule
\end{tabular}
\caption{Eligible topic pools for conditional discovery. In each seed run, three topics are randomly sampled from each pool.}
\label{tab:eligible_topics}
\end{table}

\section{Evaluation Metrics}
\label{app:evaluation_metrics}

We follow previous work~\citep{zhong2024nlparam,movva2025sparse} in evaluating synthetic experiments by matching generated hypotheses to the known ground-truth differences.
For each scenario, we repeat the experiment for 10 random seeds and report the mean and standard deviation of each metric.
We split each synthetic dataset into training, validation, and test sets in a $60{:}20{:}20$ ratio.
The validation set is used for \sae hyperparameter selection, as described in Appendix~\ref{app:hyperparameter_tuning}.

On the test set, reference annotations are determined directly by the synthetic construction.
For each generated hypothesis, we use \texttt{Qwen3-30B-A3B-Instruct-2507} to produce binary annotations on the same test texts.
Hypothesis annotation is an expensive operation.
\citet{movva2025sparse} uses \texttt{GPT-4o-mini},
and we use this open-sourced model because we can host it locally to reduce costs, and it matches \texttt{GPT-4o-mini} in annotation performances.
We internally evalutated the two models on a sample 20 hypotheses and 30 bill summaries per hypothesis, resulting in 600 text-concept pairs.
The agreement is $95.3\%$, so we use \texttt{Qwen3-30B-A3B-Instruct-2507} for hypothesis annotation throughout the experiments.
We hosted the model on two RTXA6000 GPUs with vLLM for fast inference.

Now we have one reference annotation vector (obtained from the dataset) and one generated annotation vector for each hypothesis.
We compute the correlation between every reference/generated pair, forming a $|\mathcal{H}|\times|\mathcal{H}|$ correlation matrix, and use the Hungarian algorithm to find the maximum-correlation one-to-one matching.
The \llm used for judging surface similarity is \texttt{gpt-4.1-2025-04-14}.

\begin{table*}[!htb]
\centering
\scriptsize
\renewcommand{\arraystretch}{1.3}
\setlength{\tabcolsep}{4pt}
\begin{tabular}{p{0.40\textwidth} c|p{0.40\textwidth} c}
\toprule
\lasso hypothesis & $\Delta$ & demeaned-reweighed-lasso hypothesis & $\Delta$ \\
\midrule
\multicolumn{4}{l}{\textit{Matched hypotheses}} \\
\midrule
\ncteHM{\textit{the teacher repeatedly prompts students to explain their reasoning or clarify their answers at multiple steps in multi-step fraction or mixed number problems, often after an initial incorrect or incomplete response}} &
\ncteHM{+24.7} &
\ncteHM{\textit{the teacher prompts students to explain or justify their reasoning after a partially correct or incorrect answer, and facilitates peer discussion to resolve confusion}} &
\ncteHM{+21.7} \\

\ncteHMAlt{\textit{teacher prompts students to explicitly explain or justify their procedural steps or reasoning during problem-solving, guiding them through multi-step solutions by asking follow-up questions when confusion or errors occur}} &
\ncteHMAlt{+21.9} &
\ncteHMAlt{\textit{teacher repeatedly prompts students to show all intermediate steps in their math work (e.g.\ requiring students to write down each step, not just the final answer, asking them to explain or justify each step, correcting students who do steps in their head or skip showing borrowing/regrouping, explicitly tying process visibility to grading or real-life situations)}} &
\ncteHMAlt{+16.4} \\

\ncteHM{\textit{the teacher prompts students to explicitly explain or justify their mathematical thinking while solving arithmetic problems, with multiple follow-up questions driving students to clarify errors or reasoning (rather than simply confirming answers or having students display work)}} &
\ncteHM{+23.0} &
\ncteHM{\textit{teacher asks students to explain or justify their mathematical reasoning, often following up on errors or confusions, rather than just stating answers or procedures}} &
\ncteHM{+21.0} \\

\ncteHMAlt{\textit{the teacher asks students to explicitly explain or justify their reasoning or chosen strategy for solving a problem}} &
\ncteHMAlt{+19.1} &
\ncteHMAlt{\textit{the teacher prompts students to explain or compare their solution strategies to each other}} &
\ncteHMAlt{+11.2} \\

\ncteHM{\textit{teacher prompts students to explicitly identify and correct procedural mistakes in place value or decimal operations}} &
\ncteHM{+17.2} &
\ncteHM{\textit{the teacher frequently prompts students to explain their reasoning, describe their solution process, or justify place value choices during multi-digit arithmetic problems}} &
\ncteHM{+15.3} \\

\ncteHMAlt{\textit{the teacher assigns or reminds students about their math homework specifically during lesson closure or class transitions}} &
\ncteHMAlt{+0.3} &
\ncteHMAlt{\textit{teacher assigns or discusses specific math homework or practice problems for students to complete outside of class}} &
\ncteHMAlt{+10.2} \\

\ncteLow{\textit{teacher gives procedural classroom management instructions related to cleaning up materials, lining up, or transitioning between activities}} &
\ncteLow{-7.1} &
\ncteLow{\textit{teacher directing students to physically move between locations or groups for collaborative or group work}} &
\ncteLow{-5.2} \\

\midrule
\multicolumn{4}{l}{\textit{Unique hypotheses}} \\
\midrule
\ncteLow{\textit{the teacher directs students to perform classroom management or logistical tasks, such as handing out papers, putting away materials, collecting items, or resetting groups}} &
\ncteLow{-9.7} &
& \\

\ncteLowAlt{\textit{teacher prompts students to reflect in writing (e.g.\ journals or notebooks) on what they learned, their confusions, or their problem-solving process at the end of the lesson}} &
\ncteLowAlt{-2.6} &
& \\

\ncteLow{\textit{frequent use of whole-class management language to enforce procedural compliance and orderly transitions, including directives for materials (e.g.\ whiteboards, slates), timing cues (e.g.\ minute warnings, counting down), and group routines (e.g.\ `boards up', `push your chair in'), delivered alongside academic activities}} &
\ncteLow{-6.2} &
& \\

& &
\ncteHM{\textit{teacher organizes students to work in small groups or pairs and prompts them to help and explain concepts to one another}} &
\ncteHM{+6.6} \\

& &
\ncteHMAlt{\textit{teacher provides individualized follow-up or checks in with specific students about their understanding or needs during or after whole-class instruction}} &
\ncteHMAlt{+20.6} \\

& &
\ncteLow{\textit{contains extended classroom discussions about units of weight (grams, kilograms, pounds, ounces, tons), including teacher-student exchanges involving conversions, comparisons, or real-life analogies related to weight measurement}} &
\ncteLow{-0.6} \\
\bottomrule
\end{tabular}
\caption{NCTE full hypothesis list used in the expert evaluation, with matched and unique sections in one table. Color coding: \colorbox{ncteHMLight}{green shades} indicate $\uparrow$ mid/high REMED and \colorbox{ncteLowLight}{red shades} indicate $\uparrow$ low REMED; light/lighter shades alternate by row for readability. $\Delta = \text{H/M} - \text{Low}$.}
\label{tab:ncte_full_hypotheses}
\end{table*}

\begin{table*}[!htb]
\centering
\scriptsize
\renewcommand{\arraystretch}{1.3}
\setlength{\tabcolsep}{4pt}
\begin{tabular}{p{0.40\textwidth} c|p{0.40\textwidth} c}
\toprule
\lasso hypothesis & $\Delta$ & demeaned-reweighted-lasso hypothesis & $\Delta$ \\
\midrule
\multicolumn{4}{l}{\textit{Matched hypotheses}} \\
\midrule
\congressRep{\textit{requests unanimous consent for procedural actions in the Senate, such as granting floor privileges, scheduling committee meetings, agreeing to resolutions, making technical amendments, or permitting specific actions during Senate proceedings}} &
\congressRep{+6.9} &
\congressRep{\textit{requests unanimous consent for procedural actions within the Senate or House, such as granting floor privileges, modifying amendments, scheduling votes, printing materials in the record, or controlling debate time}} &
\congressRep{+7.4} \\

\congressRepAlt{\textit{requests unanimous consent to authorize a committee or subcommittee meeting or hearing in the Senate, specifying the committee/subcommittee, date, and topic or title of the meeting}} &
\congressRepAlt{+2.9} &
\congressRepAlt{\textit{announces or authorizes official Senate committee hearings or meetings, specifying committee names, dates, times, and locations}} &
\congressRepAlt{+3.3} \\

\congressRep{\textit{gives or discusses the schedule and timing of upcoming Senate votes, bill considerations, or legislative actions (including mentioning specific days, times, or order of business)}} &
\congressRep{+7.5} &
\congressRep{\textit{discusses the scheduling and process of votes and amendments on the Senate floor, including references to specific times, sessions, and expectations for voting activity}} &
\congressRep{+4.6} \\

\congressDemAlt{\textit{criticizes government or presidential leadership for mismanagement or policy failures, particularly in areas such as war, defense, budgets, or major federal programs}} &
\congressDemAlt{-20.0} &
\congressDemAlt{\textit{criticizes government or institutional incompetence or failure to act, particularly in relation to policies, disasters, or administration}} &
\congressDemAlt{-25.1} \\

\congressDem{\textit{discusses or criticizes tax benefits, tax cuts, or loopholes that disproportionately benefit the wealthiest Americans or top income percentiles (e.g.\ top 1\%, billionaires, richest families), often contrasting with the needs of the middle or lower class}} &
\congressDem{-3.3} &
\congressDem{\textit{criticizes tax cuts or fiscal policies that disproportionately benefit wealthy individuals or families at the expense of the poor, the middle class, or public programs}} &
\congressDem{-8.5} \\

\congressDemAlt{\textit{discusses the United States national debt, including specific figures or references to the growing debt, debt ceiling, and its impact on citizens and the country}} &
\congressDemAlt{-2.8} &
\congressDemAlt{\textit{discusses the increase of the national debt or debt ceiling of the United States, with reference to specific amounts, time periods, or presidential administrations}} &
\congressDemAlt{-2.9} \\

\congressDem{\textit{criticizes proposed federal budget cuts, especially emphasizing negative impacts on socially vulnerable groups or essential services (such as veterans, students, low-income families, healthcare, or education)}} &
\congressDem{-9.6} &
\congressDem{\textit{discusses government spending, budget deficits, and the need for fiscal responsibility or limiting federal expenditures}} &
 \congressDem{-4.3} \\

\congressDemAlt{\textit{criticizes or discusses actions, policies, or leadership of the Republican Party or House Republicans}} &
\congressDemAlt{-12.1} &
\congressDemAlt{\textit{criticizes House or congressional Republicans for their policies, actions, or leadership}} &
\congressDemAlt{-11.7} \\

\midrule
\multicolumn{4}{l}{\textit{Unique hypotheses}} \\
\midrule
\congressRep{\textit{discusses progress, specific events, or achievements related to democratization and government-building in Iraq, particularly referencing Iraqi elections, voting, establishment of new governmental bodies, and the training or accomplishments of Iraqi security forces}} &
\congressRep{+0.1} &
& \\

\congressDemAlt{\textit{criticizes perceived procedural unfairness or lack of accountability within Congress or government institutions}} &
\congressDemAlt{-16.8} &
& \\

& &
\congressRep{\textit{discusses United States border security, including measures such as border fences, patrols, or enforcement against illegal crossings}} &
\congressRep{+1.6} \\

& &
\congressRepAlt{\textit{presents recent U.S. economic statistics or indicators (such as GDP growth, job creation, unemployment rate, or manufacturing output) to argue that the U.S. economy is strong or improving, often attributing this to Republican or presidential policies}} &
\congressRepAlt{+0.7} \\
\bottomrule
\end{tabular}
\caption{Congress full hypothesis list used in the expert evaluation, with matched and unique sections in one table. Color coding: \colorbox{congressRepLight}{red shades} indicate $\uparrow$ Republican and \colorbox{congressDemLight}{blue shades} indicate $\uparrow$ Democrat; light/lighter shades alternate by row for readability. $\Delta = \text{Rep.} - \text{Dem.}$}
\label{tab:congress_full_hypotheses}
\end{table*}

\section{Method Implementation Details}
\label{app:method_implementation_details}

\subsection{SAE-Based Baselines}
\label{app:sae_methods}

We compare against two global \sae feature-selection baselines.
\textbf{SAE-lasso} uses Eq.~\ref{eq:lasso} to select the top $|\mathcal{H}|$ features, using the implementation provided by \citet{movva2025sparse}.\footnote{https://github.com/rmovva/HypotheSAEs}
\textbf{SAE-separation score} ranks each feature by the mean of $y$ among the highest-activating texts versus zero-activating texts, then selects the top $|\mathcal{H}|$ features.
Both baselines use the same \sae activation matrix as the proposed methods.

\subsection{Prompts}

\begin{table*}[!htb]
\centering
\small
\begin{tcolorbox}[colback=gray!10, colframe=black, sharp corners=south, width=\textwidth, title=SAE Interpretation Prompt]
You are a machine learning researcher who has trained a neural network on a text dataset.
You are trying to understand what text features cause a specific neuron in the neural network to fire.\\

You are given two sets of SAMPLES: POSITIVE SAMPLES that strongly activate the neuron, and NEGATIVE SAMPLES from the same distribution that do not activate the neuron.
Your goal is to identify a feature that is present in the positive samples but absent in the negative samples.\\

\texttt{\{goal\}} \\

POSITIVE SAMPLES: \texttt{\{positive\_texts\}}\\

NEGATIVE SAMPLES: \texttt{\{negative\_texts\}}\\

Rules about the feature you identify:\\
The feature should be objective, focusing on concrete attributes rather than abstract concepts.
The feature should be present in the positive samples and absent in the negative samples.
Do not output a generic feature which also appears in negative samples.
The feature should be as specific as possible, while still applying to all of the positive samples.\\

Do not output anything besides the feature.
Please suggest exactly one feature, starting with \texttt{-} and surrounded by quotes.
\end{tcolorbox}

\medskip
\begin{tabular}{p{0.18\linewidth}p{0.76\linewidth}}
\toprule
\textbf{Region} & \textbf{\texttt{\{goal\}} field} \\
\midrule
Synthetic bills &
All of the texts are summaries of US legislative bills.
Features should describe a specific aspect of the bill.
For example: ``is about healthcare''; ``is about higher education.'' \\
\addlinespace
\congress &
All of the texts are excerpts of speeches from the US Congress.
Example features may include: ``mentions that the economy is strong''; ``discusses environmental issues or environmental policy.'' \\
\addlinespace
\ncte &
All texts are transcripts of $\sim$7.5-minute segments from K--12 mathematics lessons, rated on REMED (remediation of student errors and difficulties).
Example features distinguishing high-vs-low quality remediation include: ``the teacher follows up on student confusion with probing questions rather than re-explaining procedurally''; ``the teacher revoices or builds on incorrect student contributions to guide conceptual understanding''; ``students are given the opportunity to self-correct or revise their thinking.'' \\
\bottomrule
\end{tabular}
\caption{Prompt template used for the \sae interpretation step, with dataset-specific substitutions for the \texttt{\{goal\}} field.}
\label{tab:sae_interpretation_prompt}
\end{table*}

\begin{table}[!htb]
\centering
\small
\begin{tabular}{p{0.34\linewidth}p{0.34\linewidth}c}
\toprule
\textbf{Ground-truth hypothesis} & \textbf{Generated hypothesis} & \textbf{Avg. score} \\
\midrule
is about Health: Mental &
is about mental health services or policies &
1.0 \\ \addlinespace
is about Agriculture: Food Inspection \& Safety &
is about agriculture-related financial assistance &
0.5 \\ \addlinespace
is about Technology: Broadcast &
is about children's health or welfare &
0.0 \\
\bottomrule
\end{tabular}
\caption{Example matched ground-truth and generated hypotheses from the synthetic bills experiment, with averaged \llm surface-similarity judgment scores.}
\label{tab:surface_similarity_examples}
\end{table}

\subsection{LLM-Direct Baselines}
\label{app:llm_direct_sensitivity}

The LLM-direct baselines test whether covariate-aware discovery can be obtained simply by showing sampled texts to an \llm.
\textbf{llm-global} generates $|\mathcal{H}|$ hypotheses from sampled corpus texts without covariate information.
\textbf{llm-covariate} uses the same direct prompting setup but exposes covariate-profile labels.
For fair comparison, we use \texttt{gpt-4.1-2025-04-14} for hypothesis proposal in the main tables and \texttt{Qwen3-30B-A3B-Instruct-2507} for hypothesis annotation.

The LLM-direct prompts differ from the \sae interpretation prompt in Table~\ref{tab:sae_interpretation_prompt} in two ways.
First, their positive and negative examples are sampled corpus texts, rather than examples chosen by activation of a selected \sae feature.
Second, they ask the \llm to propose up to $|\mathcal{H}|$ corpus-level differences directly, rather than to name exactly one feature explaining one selected neuron.
The \textbf{llm-covariate} prompt makes one additional change relative to \textbf{llm-global}: each text is prefixed with its covariate value(s), and the model is instructed to look for patterns that are strong either globally or within a covariate stratum.

\begin{figure*}[t]
\centering
\includegraphics[width=\textwidth]{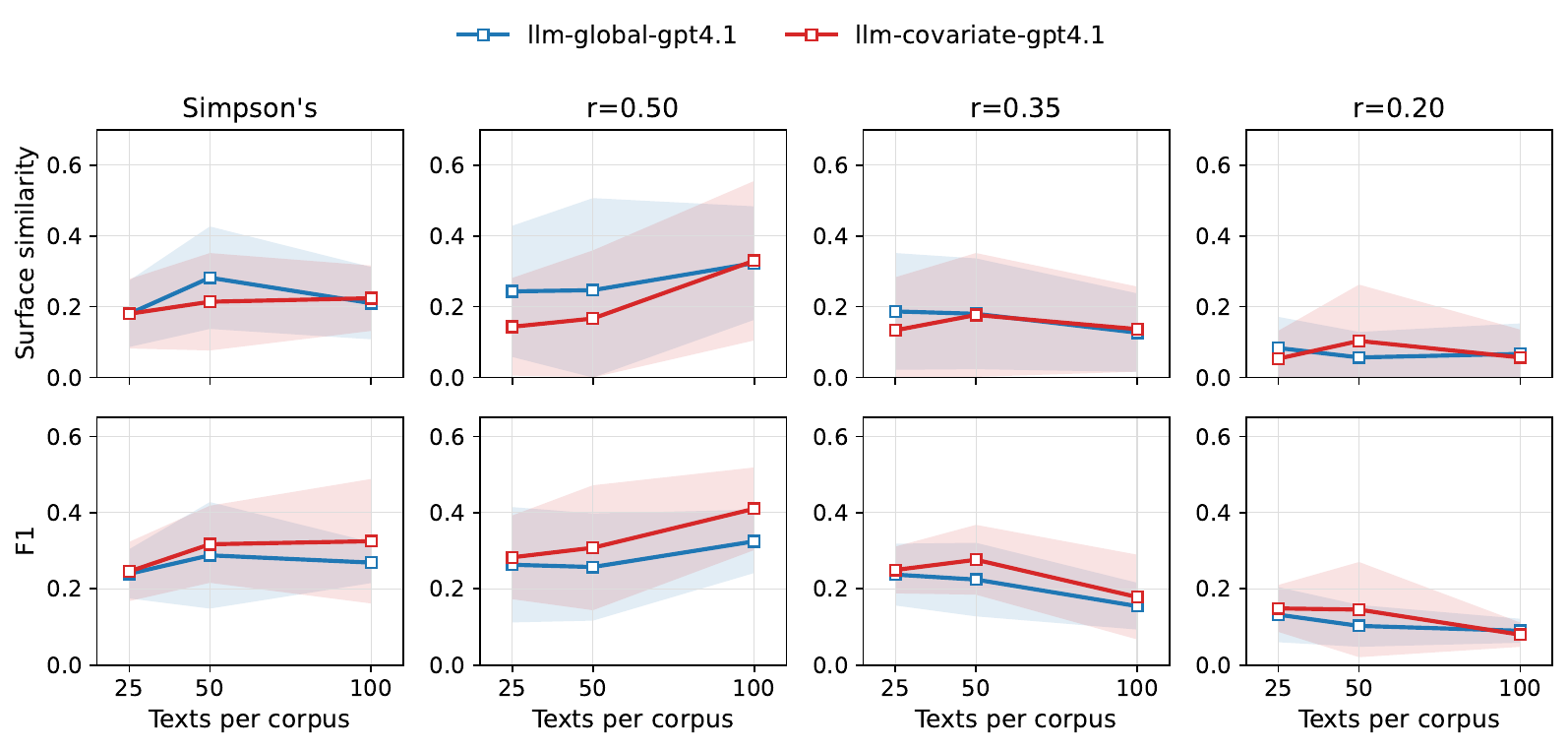}
\caption{Prompt-budget sensitivity for LLM-direct baselines using \texttt{gpt-4.1}.
Each panel varies the number of sampled texts per corpus used in the direct prompt.
Lines show seed means; shaded bands show $\pm 1$ standard deviation across seed runs.
The trends are non-monotonic: larger prompt budgets and covariate labels do not consistently improve surface similarity or F1.}
\label{fig:llm_direct_k_sensitivity_gpt41}
\end{figure*}

\begin{figure*}[!htb]
\centering
\includegraphics[width=\textwidth]{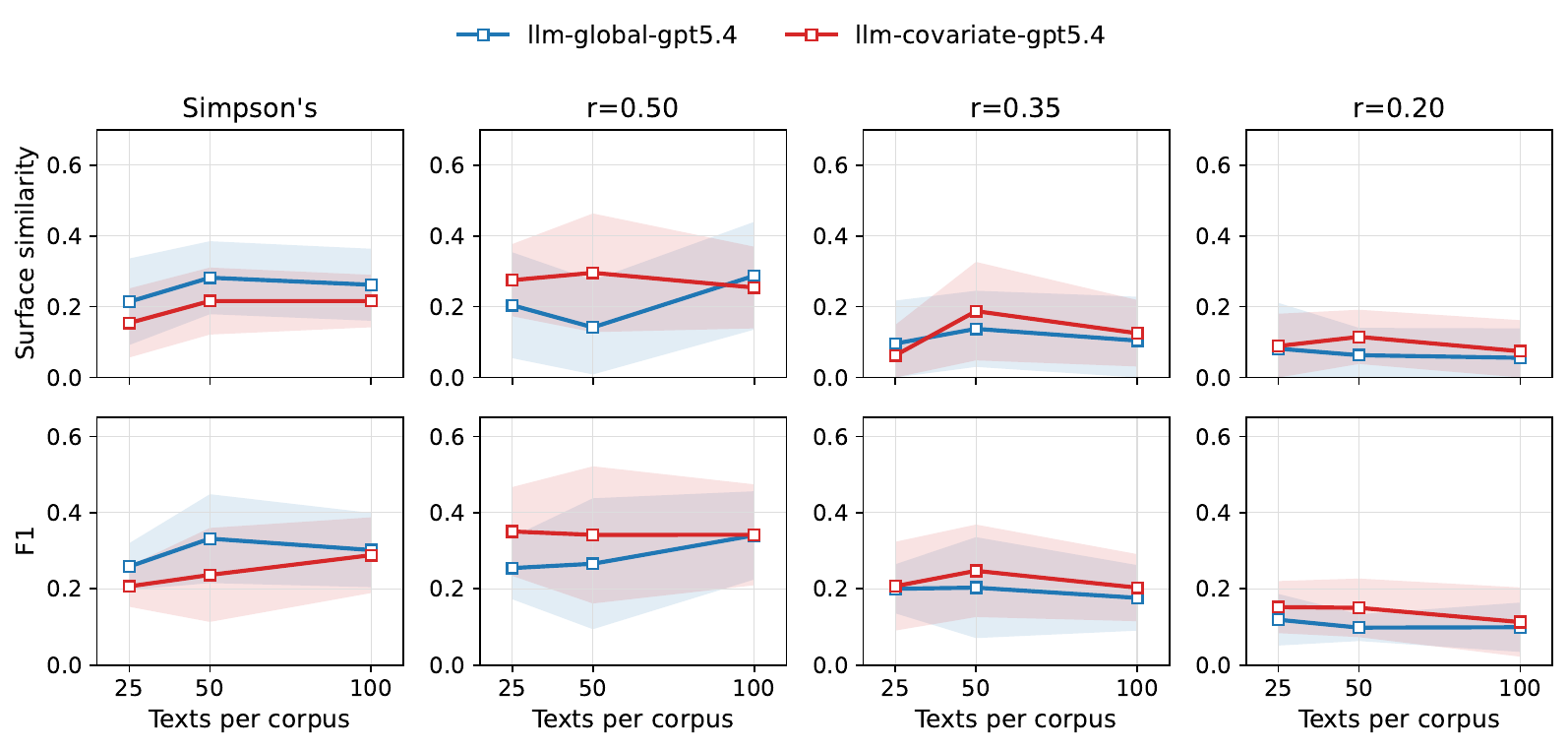}
\caption{Prompt-budget sensitivity for LLM-direct baselines using \texttt{gpt-5.4}.
Each panel varies the number of sampled texts per corpus used in the direct prompt.
Lines show seed means; shaded bands show $\pm 1$ standard deviation across seed runs.
The qualitative pattern remains non-monotonic, and covariate-labeled prompting does not consistently dominate global prompting.}
\label{fig:llm_direct_k_sensitivity_gpt54}
\end{figure*}

We swept $n_{\text{prompt}}\in\{25,50,100\}$ sampled texts per corpus and proposal models \texttt{gpt-4.1-2025-04-14} and \texttt{gpt-5.4-2026-03-05}.
Figures~\ref{fig:llm_direct_k_sensitivity_gpt41}--\ref{fig:llm_direct_k_sensitivity_gpt54} visualize the seed-level distributions across the full prompt-budget sweep.
The sweep shows that increasing the prompt budget does not consistently improve either LLM-direct baseline; covariate labels and the stronger proposal model also do not produce consistent gains.
We therefore use $n_{\text{prompt}}{=}100$ with \texttt{gpt-4.1} as a generous fixed-budget setting rather than tuning the LLM-direct baseline per condition.

\subsection{Reference Baselines}

We include two reference baselines to calibrate the synthetic evaluation.
\textbf{Oracle} is an upper reference: for each targeted ground-truth difference, $20$ texts are randomly sampled from the exact positive and negative sides of the contrast.
In Scenario 1, these are the paired $A$ and $B$ subtopics for the target social-policy topic.
In Scenario 2, these are one $\mathcal{T}_1$ subtopic versus all $\mathcal{T}_2$ subtopics combined.
These texts and their corpus labels ($A$ or $B$) are directly provided to the LLM for interpretation using the prompt in Table~\ref{tab:sae_interpretation_prompt}.
\textbf{Random} measures how much recovery can be obtained from general samples alone: $20$ training texts are randomly sampled into positive and negative halves, then passed to the same prompt.
This is not a pure lower bound, since sampled texts still contain topical cues that can receive partial credit.
We set the sample size to $20$ to be consistent with the SAE interpretation stage~\citep{movva2025sparse}.

\subsection{SAE Hyperparameter Tuning}
\label{app:hyperparameter_tuning}

We tune \sae hyperparameters on the validation split and use the selected \sae to extract feature activations for all SAE-based and proposed methods.
Following the released HypotheSAEs implementation~\footnote{https://github.com/rmovva/HypotheSAEs}, we treat the number of learned features $M$ and the per-example sparsity $K$ as the main \sae hyperparameters: $M$ controls how many concepts the model can represent, while $K$ controls how many concepts are active for each text.
The implementation recommends vanilla Top-$K$ choices of approximately $(M,K)=(64,4)$ for $\sim$1K examples, $(256,8)$ for $\sim$10K examples, and $(1024,8)$ for $\sim$100K examples, with larger $M$ yielding finer-grained concepts and $K$ typically in the range $1$--$32$.
For the synthetic bill corpora ($N=2400$) and the \ncte corpus ($\sim$10K examples), we therefore sweep a small grid around the rule-of-thumb setting:
$M\in\{128,256,512\}$ and $K\in\{4,8,16\}$.
For \congress, we instead match the larger range used for the Congressional speech experiments in~\citet{movva2025sparse}, sweeping $M\in\{1024,2048,4096\}$ and $K\in\{8,16,32\}$.
For each candidate pair, we train a vanilla Top-$K$ \sae on the training embeddings with validation embeddings for early stopping, select predictive features, and generate candidate interpretations.
On real-world datasets, the selected configuration is the one with the best validation predictive score (AUROC for binary classification, or $R^2$ for regression).
On synthetic datasets, where reference hypotheses are available, the selected configuration is the one with the highest validation surface similarity to the ground-truth hypotheses.

\section{Real-World Validation Details}
\label{app:real_world_validation}

\subsection{Matching Decisions and Full Hypothesis Lists}
\label{app:user_study_results}

Tables~\ref{tab:congress_full_hypotheses} and~\ref{tab:ncte_full_hypotheses} report the complete matched and unique lists used in the expert analysis.
Colored cells indicate direction by global prevalence difference:
Congress uses red for $\uparrow$ Republican and blue for $\uparrow$ Democrat;
NCTE uses green for $\uparrow$ mid/high REMED and red for $\uparrow$ low REMED.
All prevalence values are from the user-study annotation artifacts; $\Delta$ is the difference between the two group prevalences shown in each table.

\begin{table*}[b]
\centering
\small
\begin{tcolorbox}[colback=gray!10, colframe=black, sharp corners=south, width=\textwidth, title=Surface Similarity Prompt]
Are text\_a and text\_b similar in meaning? Respond with yes, related, or no.
\\

Here are a few examples.\\

Example 1:\\
text\_a: has a topic of protecting the environment\\
text\_b: has a topic of environmental protection and sustainability\\
output: yes
\\

Example 2:\\
text\_a: has a language of German\\
text\_b: has a language of Deutsch\\
output: yes
\\

Example 3:\\
text\_a: has a topic of the relation between political figures\\
text\_b: has a topic of international diplomacy\\
output: related
\\

Example 4:\\
text\_a: has a topic of the sports\\
text\_b: has a topic of sports team recruiting new members\\
output: related
\\

Example 5:\\
text\_a: has a named language of Korean\\
text\_b: uses archaic and poetic diction\\
output: no
\\

Example 6:\\
text\_a: has a named language of Korean\\
text\_b: has a named language of Japanese\\
output: no
\\

Example 7:\\
text\_a: describes an important 20th century historical event\\
text\_b: describes a 20th century European politician\\
output: no
\\

Target:\\
text\_a: \{text\_a\}\\
text\_b: \{text\_b\}\\
output:
\end{tcolorbox}

\caption{Prompt used for judging the surface similarity between ground-truth and inferred hypotheses. Adapted from \citet{zhong2024nlparam,movva2025sparse}.}
\label{tab:surface_similarity_prompt}
\end{table*}

\subsection{Covariate operationalization}
\label{app:cov}
For \congress, the covariate is a binary \llm annotation of whether a speech segment ``contains substantive discussion of public policy, rather than only congressional procedure.''
We use the same \congress split as \citet{movva2025sparse}: approximately 114K training, 16K validation, and 12K held-out speeches.
For \ncte, the covariate is derived from the CLASS Behavior Management score (\texttt{CLBM})~\cite{pianta2012classroom}: segments with \texttt{CLBM}$<6$ are marked as low-quality behavior management, while segments with \texttt{CLBM}$\geq6$ are marked as high-quality behavior management.
We expect to see less management language in the high-quallity segments.
This threshold marks the low and mid ranges on the CLASS scale and captures roughly 30\% of segments.
This imbalanced distribution also allows us to test the stratum imbalance scenario in a real-world dataset, which is important for \ncte as high-quality teaching samples are rare in general~\cite{kane2012gathering,xu2024promises}.
The dataset size has around 10K samples in total with 5K rated as mid/high quality \remed and 5K rated as low quality \remed. We split the dataset into train/val/testsets in a $60{:}20{:}20$ ratio.
The scores were provided by trained annotators and come with \ncte dataset.
These covariates were pre-specified before expert annotation.

\subsection{Annotation instructions}
\label{app:annotation}
Annotators reviewed method-blinded hypothesis sets from \lasso and demeaned-reweighted-lasso, shown as Set A and Set B.
They were instructed to use their domain judgment; there were no right or wrong answers.
First, they rated each hypothesis for helpfulness in understanding the outcome-group difference on a 1--5 scale (1: not helpful; 3: somewhat helpful; 5: very helpful).
They then reviewed the same hypotheses with global and covariate-stratified statistics and rated whether the covariate breakdown added interpretive value beyond the global finding, also on a 1--5 scale (1: no value; 3: some value; 5: high value).
Annotators could optionally provide free-text notes explaining their ratings.
For each dataset, we use manually curated one-to-one matched pairs and mark all remaining hypotheses as method-unique.

\subsection{Recruitment and IRB.}
\label{app:irb}
Participation was voluntary.
We invited research scholars with relevant domain expertise: computational social science researchers familiar with U.S. politics for \congress, and education researchers familiar with \ncte for the classroom-transcript study.
One \ncte annotator also had years of elementary teaching experience.
The study was reviewed by an institutional review board and determined exempt under 45 CFR 46.104(d)(2)(ii).



\end{document}